\documentclass[letterpaper]{article} 
\usepackage{aaai2026}  
\usepackage{times}  
\usepackage{helvet}  
\usepackage{courier}  
\usepackage[hyphens]{url}  
\usepackage{graphicx} 
\usepackage{natbib}  
\usepackage[table]{xcolor}
\usepackage{amsfonts}
\usepackage{amsmath}
\usepackage{mathrsfs}
\usepackage{caption} 
\usepackage{algorithm}
\usepackage{algorithmic}
\usepackage{float}
\usepackage{booktabs}
\usepackage{svg}
\usepackage{subcaption} 
\usepackage{xcolor}
\usepackage{newfloat}
\usepackage{listings}
\usepackage{mdframed}
\DeclareCaptionStyle{ruled}{labelfont=normalfont,labelsep=colon,strut=off} 
\floatstyle{ruled}
\newfloat{listing}{tb}{lst}{}
\floatname{listing}{Listing}
\title{From Ranking to Selection: A Simple but Efficient \\Dynamic Passage Selector for Retrieval Augmented Generation}

\author {
    Siyuan Meng\textsuperscript{\rm 1,2}
    Junming Liu\textsuperscript{\rm 1},
    Yirong Chen\textsuperscript{\rm 1},
    Song Mao\textsuperscript{\rm 1},
    Pinlong Cai\textsuperscript{\rm 1},
    Guohang Yan\textsuperscript{\rm 1},\\
    Botian Shi\textsuperscript{\rm 1},
    Ding Wang\textsuperscript{\rm 1}\thanks{Corresponding author.}
}
\affiliations {
    \textsuperscript{\rm 1}Shanghai Artificial Intelligence Laboratory\\
    \textsuperscript{\rm 2}East China Normal University\\
    mengsiyuancm@gmail.com, wangding@pjlab.org.cn
}

\usepackage{bibentry}

\begin{document}

\maketitle

\begin{abstract}
Retrieval-augmented generation (RAG) systems are often bottlenecked by their reranking modules, which typically score passages independently and select a \textbf{fixed Top-K} size. This approach struggles with complex multi-hop queries that require synthesizing evidence across multiple documents, creating a trade-off where small K values omit crucial information and large K values introduce noise. To address this, we introduce the \textbf{D}ynamic \textbf{P}assage \textbf{S}elector (DPS), a novel reranking framework that treats passage selection as a supervised learning problem. Unlike traditional point-wise or list-wise methods, DPS is fine-tuned to capture inter-passage dependencies and dynamically select the most relevant set of passages for generation. As a seamless plug-and-play module, DPS requires no modifications to the standard RAG pipeline. Comprehensive evaluations on five benchmarks show that DPS consistently outperforms state-of-the-art rerankers and fine-tuning methods. Notably, on the challenging MuSiQue dataset, DPS improves the F1-score by 30.06\% and 15.4\% over strong baselines like Qwen3-reranker and RankingGPT, respectively. Our results demonstrate that by enabling adaptive evidence selection, DPS substantially enhances reasoning capabilities in complex RAG scenarios.

\end{abstract}

\section{Introdction}
Information retrieval (IR)~\cite{Voorhees_1999_Natural} plays a fundamental role in many natural language processing (NLP) tasks~\cite{Hambarde_2023_Information, Wang_2024_Utilizing}, especially in retrieval-augmented generation (RAG) systems~\cite{Xia_2025_Improving, Liu_2025_VaLiK}. These systems retrieve relevant passages to support the generation process, improving the factual accuracy and reducing hallucinations~\cite{Lewis_2020_Retrieval}. In typical RAG pipelines, a fast retriever is often used to select a candidate set of passages from a large corpus, which are then re-scored by a reranker to prioritize passages that are more contextually relevant to the input query.  

Traditional reranking strategies, whether they operate at the pointwise~\cite{Nogueira_2020_Document, Liang_2023_Holistic}, pairwise~\cite{Yu_2018_WalkRanker, Qin_2024_Large}, or listwise~\cite{Thonet_2022_Listwise} level, typically follow a two-step routine: score each passage on its relevance to the query, and then pick the top K candidates as the input for the generator~\cite{Li_2023_DyRRen, Meng_2024_Ranked}. Although effective for single-hop retrieval, this approach overlooks the interdependence among passages---a crucial factor in multi-hop and complex reasoning scenarios, including QFS, long-context processing, and high-level query reasoning~\cite{GraphRAG,MemoRAG}. In many cases, the information required to answer a question is distributed across multiple sources, some of which might not be highly ranked. Consequently, traditional reranking methods face an inherent trade-off: if $K$ is too small, essential but lower-ranked evidence may be omitted; if $K$ is too large, irrelevant or distracting content may overwhelm the generator, leading to decreased performance.

\begin{figure}[t]
\hspace*{-0.05\columnwidth}
\includegraphics[width=1.05\columnwidth]{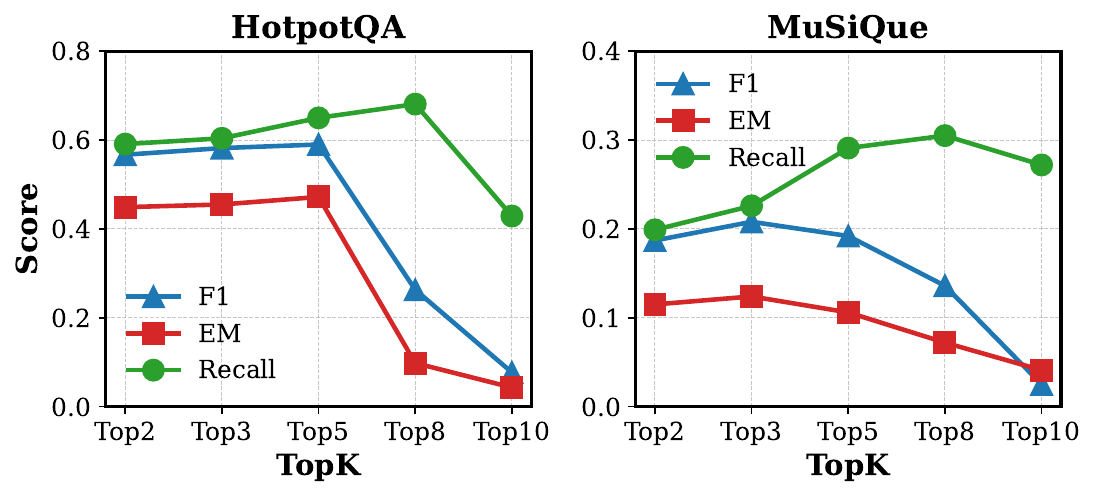}
\caption{
 Impact of Top-K passage retrieval on multi-hop QA performance. F1, EM, and Recall scores are plotted for HotpotQA and MuSiQue datasets.
}
\label{fig:intro-prelim-analysis}
\vspace{-4mm}
\end{figure}

To further illustrate this point, we conduct a preliminary analysis on two widely used multi-hop question answering benchmarks, HotpotQA~\cite{Yang_2018_Hotpotqa} and MuSiQue~\cite{Trivedi_2022}. Our findings highlight the key limitation of traditional reranking approaches: \textbf{A fixed Top-K retrieval strategy limits the ability of RAG systems to handle multi-hop and complex queries} As illustrated in Figures~\ref{fig:intro-prelim-analysis}, the optimal value of $K$ is highly query-dependent, $K{=}5$ for HotpotQA and $K{=}3$ for MuSiQue. Increasing $K$ from 2 to 8 improves recall by up to $13.21\%$ and $53.26\%$ on HotpotQA and MuSiQue, respectively, demonstrating the benefit of incorporating more contextual information. However, the F1 score drops sharply beyond $K{=5}$ on HotpotQA and $K{=3}$ on MuSiQue, indicating diminishing or even negative returns as irrelevant content begins to dilute the useful evidence.

\begin{figure*}
\centering
\includegraphics[width=0.95\textwidth]{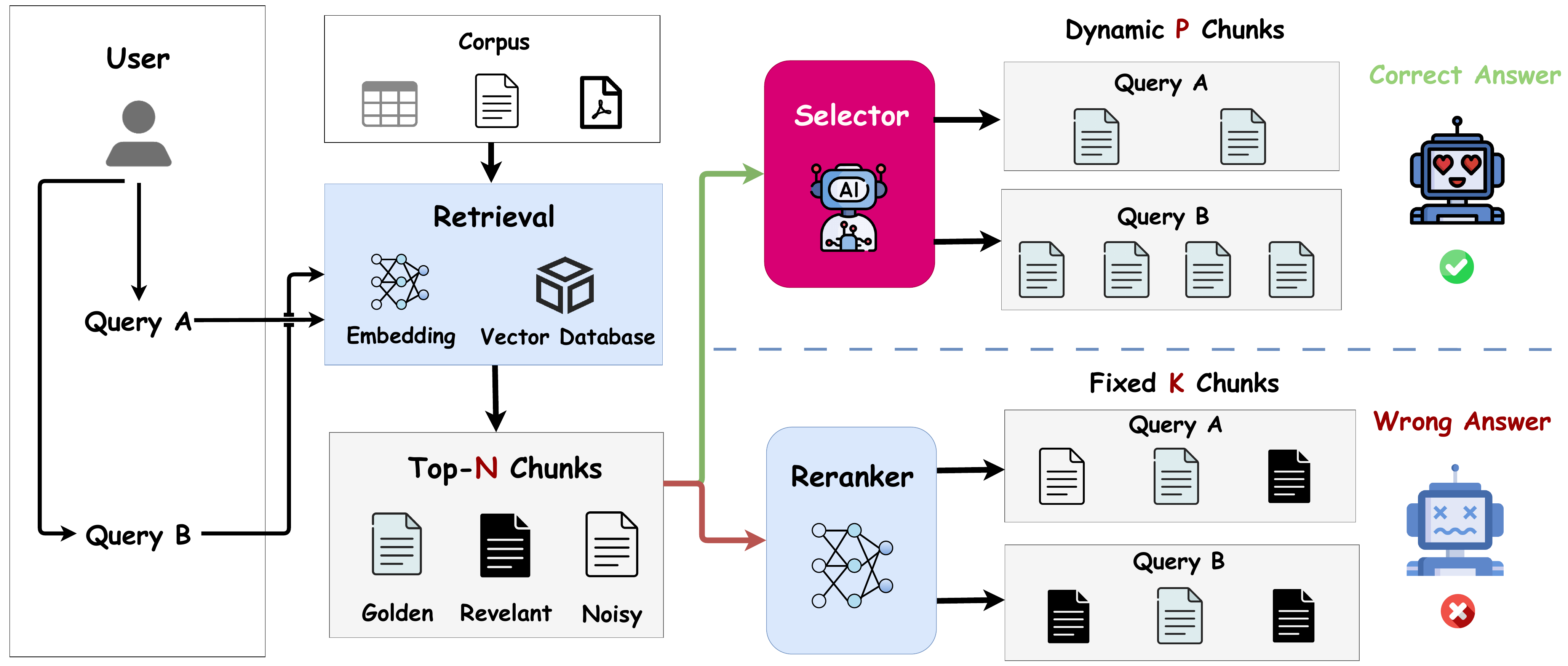}
\caption{Comparison between fixed-\textit{K} reranking and the proposed dynamic passage selector DPS. Given a user query, a retrieval module fetches Top-\textit{N} candidate passages from a large corpus. DPS adaptively selects a minimal yet sufficient subset of passages, enabling more accurate multi-hop reasoning and generation. 
}
\label{fig:intro_ranker_vs_selector}
\end{figure*}

To address this problem, we propose to reframe reranking as a passage selection learning problem, where large language models are fine-tuned through supervised learning (SFT) to identify the minimal set of passages that collectively support answering a query. Rather than assigning scalar relevance scores to each passage, the proposed framework,\textbf{ Dynamic Passage Selector (DPS)}, learns to reason about the combinatorial relevance. This formulation allows the model to explicitly capture the inter-passage dependencies critical for multi-hop question answering, understanding when a single passage would suffice or when multiple passages must be selected together.

As illustrated in Figure~\ref{fig:intro_ranker_vs_selector}, the number of passages required to answer the query varies between query A and query B. Traditional reranking approaches retrieve a fixed Top-K set of passages, which introduces redundant information that may compromise answer quality or lead to incorrect responses. In contrast, DPS dynamically selects the minimal yet sufficient subset of passages based on query-specific needs, thereby ensuring the quality of generated content. Additionally, recent methods attempt to align retrieval and ranking in RAG pipelines~\cite{Li_2024_Learning, Zhang_2025_Killing, Zuccon_2025_R2LLMs}, they often rely on costly multi-stage API calls or require large-scale fine-tuning. By contrast, the proposed DPS can be seen as a \textbf{plug-and-play} component that integrates seamlessly into existing RAG workflows without fine-tuning process or pipeline modifications.

We validate our approach on a range of question answering benchmarks with varying levels of difficulty, including three popular multi-hop question answer datasets HotpotQA~\cite{Yang_2018_Hotpotqa}, MuSiQue~\cite{Trivedi_2022},  2WikiMQA~\cite{Ho_2020_2WikiMultihopQA}, as well as two out-of-domain datasets Legal and CS~\cite{MemoRAG}. 
Experimental results show that our dynamic passage selector consistently outperforms traditional point-wise rerankers and state-of-the-art LLM-based list-wise rerankers, especially on complex multi-hop queries. The results demonstrate that DPS can dynamically identify contextually relevant passages, surpassing the limitations of fixed top-K retrieval. By more effectively aligning retrieval with generation, DPS leads to more robust end-to-end performance. Our main contributions are as follows:

\begin{itemize}
    \item  To the best of our knowledge, this is the first work that identifies and addresses a critical limitation in RAG systems: the fixed Top-K retrieval strategy severely impairs RAG system's performance on multi-hop and complex queries.
    \item We treat reranking as a passage selection problem, fine-tune LLMs to dynamically select a minimal set of sufficient passages by modeling their interdependencies, enabling accurate retrieval for multi-hop and complex query, while remaining fully plug-and-play with existing RAG systems.
    \item Through extensive evaluations across five diverse question answering benchmarks, we demonstrate that DPS achieves state-of-the-art performance on both challenging multi-hop and out-of-domain datasets (e.g., 30.06\% F1 increase over Qwen3-reranker on MuSiQue), validating its capabilities and robustness for RAG systems.
\end{itemize}    
\section{Related Work}
\subsection{Traditional and Neural Reranking}
Text ranking is a core task in IR, where the goal is to identify and rank the most relevant documents for a user query~\cite{Yates_2021_Pretrained}. Modern IR systems often adopt a two-stage architecture: a first-stage retriever (e.g., BM25~\cite{Robertson_1995_BM25} or dense retrievers~\cite{Kang_2025_Distribution, Ma_2025_Task}) retrieves a broad set of candidate documents, and a reranker refines this list to ensure the top-ranked documents are semantically relevant~\cite{Lin_2021_Pyserini}. Rerankers model fine-grained semantic interactions and significantly boost performance over first-stage retrieval. Recent reranking approaches predominantly rely on cross-encoder architectures, where the query and each document are jointly encoded to compute semantic similarity~\cite{Gao_2025_LLM4Rerank}. Open-source models like bge-reranker-large~\cite{Chen_2024_BGE} offer strong performance and allow flexible fine-tuning across domains, rivaling proprietary models such as Cohere-reranker~\cite{cohere_reranker}. Beyond cross-encoders, T5-based models such as MonoT5~\cite{MonoT5} and RankT5~\cite{Zhang_2023_RankT5} reformulate ranking as a sequence generation task, showing that encoder-only variants may suffice for many reranking problems. However, these models typically treat each document independently—following pointwise or pairwise ranking paradigms—which limits their ability to capture inter-document dependencies, especially for complex reasoning tasks.

\subsection{LLM-based Approaches}
With the rise of LLMs, new efforts aim to leverage their generative abilities for listwise ranking~\cite{Zhuang_2024_A, Podolak_2025_Beyond}. For example, Listwise Ranking via Language Models prompts GPT-3 to rank documents by generating an ordered list of document IDs~\cite{ma2023zero}, allowing global relevance reasoning over all candidates. Despite the promise of LLMs in reranking, Liu et al. observe that vanilla prompting often fails, particularly in smaller-scale LLMs, due to the misalignment between next-token prediction objectives and ranking tasks~\cite{Qin_2024_Large}. Their proposed Pairwise Ranking Prompting (PRP) combines prompting with scoring, improving ranking performance even for open-source models like LLaMA and Mistral. Similarly, RankingGPT~\cite{zhang2023rankinggpt} addresses this misalignment by introducing a two-stage training framework: query generation pretraining followed by supervised fine-tuning with hard negatives. Their results emphasize that supervised fine-tuning is critical for aligning LLM behavior with ranking objectives. Although these works explore listwise and generation-based paradigms, they often still operate at the level of individual documents and do not explicitly model the combinatorial relationships needed for multi-hop reasoning. Our approach builds on this line of work but moves beyond scalar relevance estimation by treating reranking as a document selection problem guided by reasoning over document sets.
\section{Methodology}
\begin{figure*}[t]
\centering
\includegraphics[width=1.0\textwidth]{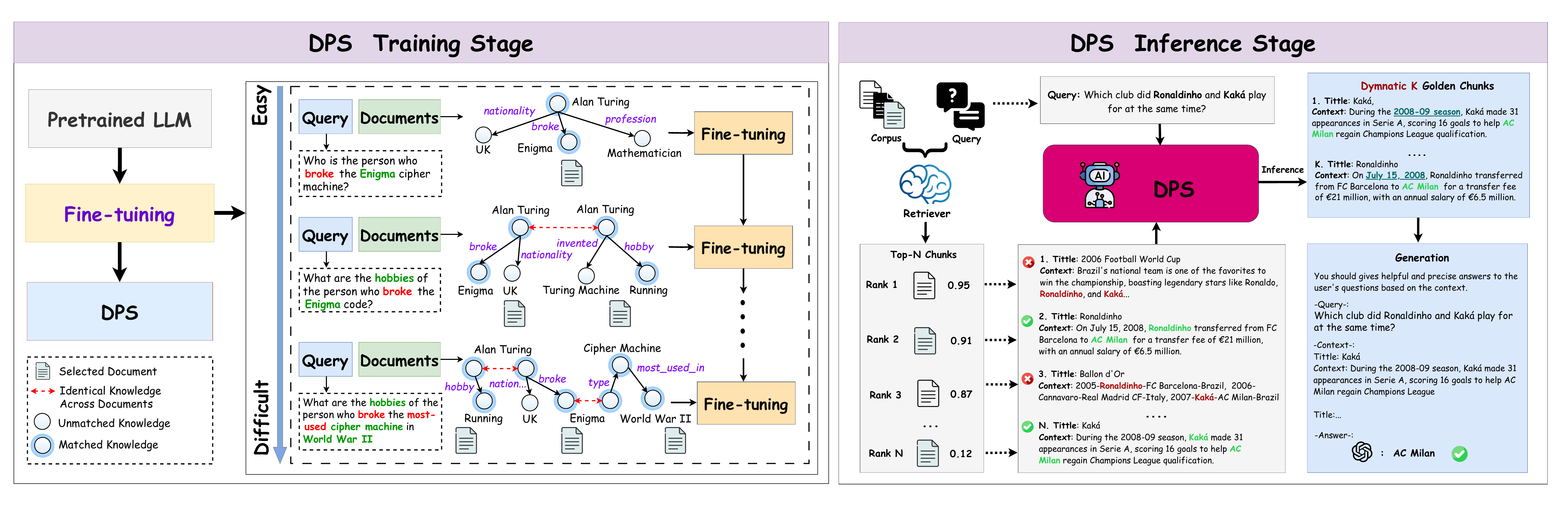}
\caption{During the training phase, we treat single-hop question answering data as easy samples, while multi-hop question answering and complex queries as difficult samples. During the inference phase, DPS adaptively selects K passages from the N passages retrieved by fast retriever to provide to the generator.}
\label{fig:overview}
\end{figure*}

This section presents the core methodology of DPS.
As shown in Figure \ref{fig:overview}, the DPS framework comprises two primary stages: offline supervised fine-tuning and online inference. During training, DPS formulates passage selection as a conditional sequence modeling task, enabling the model to learn the interdependencies among candidate passages and to identify the most informative subset. At inference time, DPS functions as a lightweight, plug-and-play component that can be readily integrated into existing RAG pipelines without requiring architectural modifications and additional training.

\subsection{Problem Formulation}
Given a natural language query \( q \in \mathbb{Q} \) and a large text corpus \( \mathcal{C} = \{ c_1, c_2, \ldots \} \), a fast retriever module retrieves a candidate set of passages:

\[
\mathcal{P} = \{p_1, p_2, \ldots, p_n\} \subseteq \mathcal{C},
\]
where each \( p_i \) denotes a retrieved passage relevant to the query \( q \), and \( n \) is the number of retrieved candidates (typically large, e.g., Top-100).

Traditional rerankers assign an independent relevance score \( s_i = f(q, p_i) \) to each passage using a scoring function \( f(\cdot) \), and select the top-\(K\) passages based on these scores:

\[
\mathcal{S}_{\text{top-}K} = \{ p_i \mid i \in \arg\text{top}_{|S|=K} s_i \},
\]
where \( \mathcal{S}_{\text{top-}K} \subseteq \mathcal{P} \) denotes the fixed-size selected subset and \( K \) is a predefined constant.

However, this standard approach has several key limitations that we discussed before. First, the value of \( K \) is query-agnostic and fails to adapt to varying levels of query complexity. For instance, single-hop questions may require only one passage, whereas complex multi-hop questions may demand a larger and more diverse evidence set. Second, the scoring function \( f(q, p_i) \) in traditional methods lacks explicit modeling of subset-level reasoning, and consequently struggles to identify combinations of passages that jointly offer comprehensive support for answer generation.

In contrast, our proposed DPS learns to select a \textit{query-specific} subset of passages of variable size, optimized to support accurate answer generation with minimal redundancy. The selection process is formulated as a sequence prediction task, which we describe in the following section.

\subsection{Dynamic Passage Selection}

We reformulate passage selection as a structured prediction problem over variable-length subsets of retrieved candidates. The training and inference process of DPS is illustrated in Figure ~\ref{fig:overview}. Given a natural language query \( q \in \mathbb{Q} \) and a retrieved set of passages \( \mathcal{P} = \{p_1, p_2, \ldots, p_n\} \), the goal is to select a minimal yet sufficient subset \( S \subseteq \mathcal{P} \) that supports accurate answer generation. We model this as learning a conditional probability distribution over subsets:

\[
P_\theta(S \mid q, \mathcal{P}), \quad \text{where } S \subseteq \mathcal{P},
\]
where \( \theta \) denotes model parameters, and the size \( |S| = k \) is not fixed, but instead determined dynamically depending on the complexity of the input query \( q \).

To enable efficient modeling, we represent the subset \( S \) as an ordered sequence of indices \( (i_1, i_2, \ldots, i_k) \), where \( i_j \in \{1, \ldots, n\} \) corresponds to the position of passage \( p_{i_j} \in \mathcal{P} \). The distribution is then factorized in an autoregressive manner:

\[
P_\theta(S \mid q, \mathcal{P}) = \prod_{j=1}^k P_\theta(i_j \mid q, \mathcal{P}, i_{<j}),
\]
where \( i_{<j} = (i_1, \ldots, i_{j-1}) \) denotes the previously selected indices. 
This formulation has two key benefits. First, each decision \( P(i_j \mid q, \mathcal{P}, i_{<j}) \) is explicitly conditioned on the query \( q \), ensuring alignment with the information need. Secondly, conditioning on \( i_{<j} \) captures redundancy or complementarity among passages, promoting diversity and coherence in the final subset. This autoregressive structure allows the model to terminate generation once a sufficient set has been selected, thus avoiding the need to predefine the subset size \( k \).

We implement \( P_\theta \) using a decoder-only large language model fine-tuned for passage index prediction. 
The model input is constructed by concatenating the query and the retrieved passages, where each passage \( p_i \) is prepended with a unique index token:

\[
\text{Input} = \text{Query: } q \ \Vert\ \text{Passages: } [1]p_1\ [2]p_2\ \ldots\ [n]p_n,
\]
where \([i]\) denotes the index token associated with passage \( p_i \). The model is trained to generate the output sequence:

\[
\text{Output} = i_1, i_2, \ldots, i_k.
\]
where each \( i_j \) indicates the index of a selected passage. This formulation enables the model to directly predict informative subsets by leveraging the LLM’s strong sequence modeling and in-context reasoning capabilities, all without requiring architectural changes.

\subsection{Supervised Fine-tuning}

To effectively train the model, we construct a supervised dataset containing tuples \( (q, \mathcal{P}, S^*) \), where:
\begin{itemize}
    \item \( q \) is the input query,
    \item \( \mathcal{P} = \{p_1, \ldots, p_n\} \) is the retrieved candidate passage set,
    \item \( S^* = (i_1^*, \ldots, i_k^*) \) is the ground-truth ordered subset of indices sufficient to answer \( q \).
\end{itemize}

Given a training dataset \( \mathcal{D} = \{(q^{(m)}, \mathcal{P}^{(m)}, S^{*(m)})\}_{m=1}^M \), we minimize the sequence-level cross-entropy loss:

\[
\mathcal{L}(\theta) = - \frac{1}{M} \sum_{m=1}^M \sum_{j=1}^{k_m} \log P_\theta \bigl(i_j^{*(m)} \mid q^{(m)}, \mathcal{P}^{(m)}, i_{<j}^{*(m)}\bigr),
\]

where \(m\) denotes the \( m \)-th training instance in the dataset of size \( M \), \( i_{<j}^{*(m)} = (i_1^{*(m)}, \ldots, i_{j-1}^{*(m)}) \) denotes the previously selected indices. 

This training process encourages the model to: 1) predict a minimal set of passages that jointly cover the required evidence; 2) capture passage-level dependencies and avoid redundancy; 3) adapt to variable query complexity by generating dynamic-length outputs.

\subsection{Inference Strategy}
At inference time, DPS receives a query \( q \) and a retrieved candidate set \( \mathcal{P} = \{p_1, \ldots, p_n\} \) obtained from a fast retriever (e.g., BM25 or embedding model). The model performs autoregressive generation over passage indices to construct a variable-sized subset \( S = (i_1, i_2, \ldots, i_k) \), where each index \( i_j \in \{1, \ldots, n\} \) denotes the position of a selected passage in \( \mathcal{P} \).

The generation proceeds step-by-step, conditioned on the query and previously selected indices:

\[
i_j \sim P_\theta(i_j \mid q, \mathcal{P}, i_{<j}), \quad \text{for } j = 1, 2, \ldots, k.
\]
The final output is a dynamically determined subset \( S \subseteq \mathcal{P} \), which is concatenated and used as the context input to the downstream generator. Importantly, this process requires no changes to the generator architecture or training pipeline, making DPS a plug-and-play component compatible with standard RAG systems.

\subsection{Fine-tuning Data Construction}
Fine-tuning the DPS model requires a corpus of accurately labeled QA data, including high-quality examples of both single-hop and multi-hop/complex queries, each paired with its set of supporting passages (Detailed statistics are provided in the appendix.). Such data is crucial for training the model to capture passage-level dependencies between a query and the evidence needed to answer it.
To this end, we perform comprehensive data collection (shown as Table~\ref{tab:train_data}) from two primary sources: existing labeled QA datasets and synthetic generation.

We collect a diverse and high-quality set of fine-tuning data from established benchmarks.For multi-hop QA, we incorporate data from HotpotQA and MuSiQue, which source their corpora from Wikipedia. These datasets provide examples where multiple documents are needed to formulate an answer.
For single-hop QA, to bolster the model's single-hop reasoning capabilities and diversify our data distribution, we curated 100k question-passage pairs from the MS MARCO~\cite{MSMARCO} dataset. The vast majority of these (approximately 95\%) are single-evidence queries, requiring only one passage for a complete answer.

We generate synthetic data to mitigate the shortage of long context and high-level queries. We use GPT-4o to generate synthetic data, and the prompt is “You are a curious AI assistant, please generate one specific and valuable question based on the following passages. The generated question should revolve around the core content of this text. Note that the generated questions need to be answered by combining information from all the passages:”.
\begin{table}[t]
\centering
\begin{tabular}{@{}ccc@{}}
\toprule
Data Source    & Type                                                              & Size \\ \midrule
MS MARCO       & \begin{tabular}[c]{@{}c@{}}single-hop and multi-hop\end{tabular}    & 100k \\
HotpotQA       & multi-hop                                                         & 90k  \\
MuSiQue        & multi-hop                                                         & 20k  \\
Synthetic data & \begin{tabular}[c]{@{}c@{}}long context and high-level\end{tabular} & 5k   \\ \bottomrule
\end{tabular}
\caption{Specification of training data.}
\label{tab:train_data}
\end{table}

\begin{table*}[t]
\centering
\setlength{\tabcolsep}{9pt}
\renewcommand{\arraystretch}{1.1}
\begin{tabular}{llllllllclc}
\hline
Method/Dataset  & \multicolumn{2}{c}{HotpotQA}                    & \multicolumn{2}{c}{Musique}                     & \multicolumn{2}{c}{2WikiMQA}                     & \multicolumn{2}{c}{Legal}   & \multicolumn{2}{c}{CS}      \\ \hline
Metric          & \multicolumn{1}{c}{F1} & \multicolumn{1}{c}{EM} & \multicolumn{1}{c}{F1} & \multicolumn{1}{c}{EM} & \multicolumn{1}{c}{F1} & \multicolumn{1}{c}{EM} & \multicolumn{1}{c}{F1} & EM & \multicolumn{1}{c}{F1} & EM \\ \hline
\rowcolor{gray!10}
\multicolumn{11}{c}{w/o RAG}                                                                                                                                                                                                      \\ \hline
GPT-4o          & 41.72                  & 30.74                  & 14.07                  & 05.97                  & 34.31                  & 28.42                  & 18.27                  & -  & 20.51                  & -  \\
Gemini1.5-flash & 29.74                  & 22.75                  & 07.59                  & 03.22                  & 28.11                  & 25.71                  & 16.93                  & -  & 17.76                  & -  \\
Qwen2.5-72B     & 23.20                  & 18.47                  & 03.32                  & 01.53                  & 14.42                  & 12.91                  & 14.00                  & -  & 14.60                  & -  \\
Llama3.1-70B    & 40.06                  & 28.66                  & 13.75                  & 04.56                  & 31.04                  & 25.39                  & 17.36                  & -  & 17.86                  & -  \\ \hline
\rowcolor{gray!10}
\multicolumn{11}{c}{Fine-tuning Method}                                                                                                                                                                                           \\ \hline
DPA-RAG         & 57.39                  & 46.72                  & 20.45                  & 17.36                  & 39.82                  & 39.22                  & \multicolumn{1}{c}{-}  & -  & \multicolumn{1}{c}{-}  & -  \\
RankRAG         & 46.70                  & 35.30                  & \multicolumn{1}{c}{-}  & \multicolumn{1}{c}{-}  & 36.90                  & 31.40                  & \multicolumn{1}{c}{-}  & -  & \multicolumn{1}{c}{-}  & -  \\ \hline
\rowcolor{gray!10}
\multicolumn{11}{c}{Reranker}                                                                                                                                                                                                     \\ \hline
BGE-reranker-v2-m3    & 58.17                  & 45.46                  & 19.14                  & 10.49                  & 23.44                  & 19.35                  & 23.92                  & -  & 19.47                  & -  \\
Qwen3-reranker-8B  & 60.45                  & 47.37                  & 22.08                  & 13.10                  & 32.20                  & 28.71                  & 22.10                  & -  & 19.94                  & -  \\
Mono-T5  & 55.31                  & 42.11                  & 18.23                  & 11.14                  & 21.93                  & 19.21                  & 20.10                  & -  & 19.01                  & -  \\
RankingGPT         & 61.12                  & 48.10                  & 24.98                  & 14.97                  & 36.39                  & 32.17                  & 21.48                  & -  & 18.32                  & -  \\
RankLlama       & 54.28                  & 42.03                  & 19.61                  & 11.48                  & 27.16                  & 23.58                  & 22.46                  & -  & 18.44                  & -  \\
RankZephyr      & 53.15                  & 41.49                  & 19.73                  & 11.90                  & 34.15                  & 30.11                  & 22.37                  & -  & 18.59                  & -  \\
RankVicuna      & 52.81                  & 40.12                  & 19.05                  & 11.72                  & 33.25                  & 29.87                  & 21.58                  & -  & 18.27                  & -  \\ \hline
DPS             & \textbf{66.34}         & \textbf{53.28}         & \textbf{28.85}         & \textbf{19.57}         & \textbf{46.13}         & \textbf{41.47}         & \textbf{28.72}         & -  & \textbf{29.37}         & -  \\ \hline 
\end{tabular}
\caption{ Experimental results on five benchmarks, where we highlight the best results in \textbf{bold}. We compute the average EM and F1 on three multi-hop QA datasets, and average F1 on CS and Legal. }
\label{table:main}
\end{table*}

\section{Experiment}
\subsection{Experitmental Setups}
\subsubsection{Datasets.}
We evaluate DPS on five widely used question answering benchmarks covering both multi-hop and domain-specific reasoning. HotpotQA~\cite{Yang_2018_Hotpotqa} requires concatenating evidence from two supporting Wikipedia paragraphs to answer complex questions. MuSiQue~\cite{Trivedi_2022} similarly provides multi-hop queries with paragraph‑level annotations, but emphasizes compositional reasoning chains. 2WikiMQA~\cite{Ho_2020_2WikiMultihopQA} comprises factoid questions that may span one or two Wikipedia articles. To evaluate long context and high-level query, we include Legal~\cite{MemoRAG}, a collection of legal-domain questions with specialized terminology, and CS~\cite{MemoRAG}, which focuses on computer science and programming knowledge. 

\subsubsection{Baselines.}
We compare against popular pre-trained models in zero‑shot mode (GPT‑4o, Gemini-1.5‑flash, Qwen2.5‑72B, and Llama3.1‑70B) to measure the benefit of retrieval augmentation. In the fine‑tuned RAG category, we include DPA‑RAG~\cite{DPA-RAG} and RankRAG~\cite{Rank-RAG}, both of which jointly optimize retriever and generator components using passage‑level supervision and contrastive or distillation objectives. For reranker-augmented methods, we evaluated seven standard rerankers—BGE-reranker-v2-m3\cite{Chen_2024_BGE}, Qwen3-reranker-8B\cite{qwen3-reranker}, Mono-T5\cite{MonoT5}, RankingGPT~\cite{zhang2023rankinggpt}, RankLlama~\cite{rankllama}, RankZephyr\cite{rankze} and RankVicuna~\cite{rankv}—that apply a pre-trained reranking model over fast retriever candidates before feeding the top subset into an LLM.

\subsubsection{Evaluation.}
Compared to previous reranking models that focus on retrieval-based metrics, we place greater emphasis on whether the retrieved passages can improve the quality of the generated content.
All methods are assessed using token‑level F1 and Exact Match (EM) metrics on multi-hop QA dataset. For the CS and Legal datasets, the ground-truth consists of long-form answers, so only F1 is used as the evaluation metric. The experiments are conducted on a single NVIDIA A100 GPU with a fixed test batch size and temperature. Each configuration is run three times with different random seeds. During evaluation, for each dataset, we vary the top-k parameter in baseline rerankers from 1 to 8 to measure performance across different evidence sizes, and find K=5 is the optimal settings for most rerankers on five datasets.

\subsubsection{Implementation.}
We fine-tune DPS on Qwen2.5-7B using LoRA adapters on 4 NVIDIA A100 GPUs. 
We set training epoch $I=1$ and use the DeepSpeed ZeRO
strategy during training, with learning rates of $1\mathrm{e}{-4}$ and a warm ratio of 0.05. To ensure fairness, we use the same vector retrieval method to retrieve 30 documents from the identical corpus, serving as the candidate set for both the reranker and DPS. We use one of the most advanced text-encoding models, BGE-M3~\cite{M3}, as the fast retriever across all reranker methods. We use Qwen2.5-7B as the generator (backbone LLMs) for all the baselines in main results. The detail implementation of baseliens and our source code can be found in the Appendix.

\begin{table*}[t]
\centering
\setlength{\tabcolsep}{12pt}
\begin{tabular}{@{}ccccccccccc@{}}
\toprule
Method    & \multicolumn{2}{c}{HotpotQA}    & \multicolumn{2}{c}{Musique}     & \multicolumn{2}{c}{2WikiMQA}     & \multicolumn{2}{c}{Legal} & \multicolumn{2}{c}{CS} \\ \midrule
Metric    & F1             & EM             & F1             & EM             & F1             & EM             & F1                & EM    & F1               & EM  \\ \midrule
-SFT      & 44.41          & 34.87          & 14.26          & 08.71          & 27.47          & 24.78          & 19.66             & -     & 20.65            & -   \\
-MS  & 65.77          & 52.70          & 27.53          & 18.49          & 45.83          & 41.33          & 22.15             & -     & 23.62            & -   \\
-Multihop & 47.00          & 36.53          & 14.65          & 08.87          & 27.04          & 23.58          & 28.51             & -     & \textbf{29.81}   & -   \\
-Synthetic & 64.21          & 52.05          & 28.02          & 18.67          & 46.02          & 40.39          & 28.20             & -     & 29.21   & -   \\ \midrule
DPS       & \textbf{66.34} & \textbf{53.28} & \textbf{28.85} & \textbf{19.57} & \textbf{46.13} & \textbf{41.47} & \textbf{28.72}    & -     & 29.37            & -   \\ \bottomrule
\end{tabular}
\caption{Ablation study on the effects of different training components on DPS performance across multiple datasets.}
\label{tab:ablation}
\end{table*}

\begin{table*}[t]
\setlength{\tabcolsep}{10pt}
\begin{tabular}{@{}ccccccccccc@{}}
\toprule
Method              & \multicolumn{2}{c}{HotpotQA}    & \multicolumn{2}{c}{Musique}     & \multicolumn{2}{c}{2WikiMQA}     & \multicolumn{2}{c}{Legal} & \multicolumn{2}{c}{CS} \\ \midrule
Metric              & F1             & EM             & F1             & EM             & F1             & EM             & F1                & EM    & F1               & EM  \\ \midrule
Qwen2.5-7B          & 15.53          & 12.69          & 01.05          & 00.37          & 13.38          & 13.03          & 16.38             & -     & 18.36            & -   \\
Qwen2.5-72B         & 23.20          & 18.47          & 03.32          & 01.53          & 14.42          & 12.91          & 14.00             & -     & 14.60            & -   \\
Llama3.1-8B         & 21.09          & 12.90          & 05.83          & 01.00          & 15.13          & 11.69          & 17.91             & -     & 18.98            & -   \\
Gemini1.5-flash     & 29.74          & 22.75          & 07.59          & 03.22          & 28.11          & 25.71          & 16.93             & -     & 17.76            & -   \\
GPT-4o              & 41.72          & 30.74          & 14.07          & 05.97          & 34.31          & 28.42          & 18.27             & -     & 20.51            & -   \\ \midrule
\rowcolor{gray!10}
\multicolumn{11}{c}{DPS}                                                                                                                                                       \\ \midrule
DPS+Qwen2.5-7B      & 66.34          & 53.28          & 28.85          & 19.57          & 46.13          & 41.47          & 28.72             & -     & 29.37            & -   \\
DPS+Qwen2.5-14B     & 62.91          & 46.39          & 26.62          & 17.33          & 43.00          & 33.29          & 29.08             & -     & 29.28            & -   \\
DPS+Qwen2.5-72B     & \textbf{69.02} & \textbf{54.14} & 30.03          & 19.53          & 47.18          & \textbf{42.03} & 23.62             & -     & 22.22            & -   \\
DPS+Llama3.1-8B     & 62.80          & 49.09          & 25.91          & 15.96          & 33.44          & 24.54          & 29.68             & -     & 31.62            & -   \\
DPS+Genemi1.5-flash & 66.92          & 52.11          & 31.14          & 17.16          & 43.34          & 37.73          & \textbf{33.13}    & -     & \textbf{38.93}   & -   \\
DPS+GPT-4o          & 68.65          & 53.07          & \textbf{32.40} & \textbf{21.06} & \textbf{46.23} & 41.78          & 30.67             &       & 32.27            & -   \\ \bottomrule
\end{tabular}
\caption{Performance of DPS combined with various backbone LLMs with different sizes.}
\label{tab:further_exp}
\end{table*}

\subsection{Main Results}
As shown in Table~\ref{table:main}, DPS achieves SOTA performance across all five QA benchmarks, outperforming both fine-tuned generation baselines (e.g., DPA-RAG, RankRAG) and recent LLM-based reranking models (e.g., RankingGPT, RankLlama). 
On average, DPS improves F1 scores by 8.1\% and EM scores by 7.9\% over strong rerankers. 
On multi-hop datasets such as HotpotQA, MuSiQue, and 2WikiMQA, DPS achieves consistent improvements over strong reranking-based baselines. Specifically, DPS improves F1 by 5.2 points on HotpotQA (66.34 vs. 61.12 against RankingGPT) and 3.87 points on MuSiQue (28.85 vs. 24.98). On 2WikiMQA, DPS reaches 46.13 F1, surpassing RankingGPT by 9.74 points. 

Beyond standard QA tasks, DPS shows strong generalization to domain-specific datasets such as Legal and CS. Despite the absence of domain-specific fine-tuning, DPS outperforms DPA-RAG and RankingGPT by up to 6.3\% in F1 on the Legal and CS domains. This demonstrates DPS's strong capability in handling long contexts and high-level queries. Moreover, this also suggests that dynamic passage selection is more adaptable to unseen tasks and domains compared to rigid Top-K reranking strategies. 

These results confirm that the passage selection approach generalizes well across both standard and multi-hop question answering tasks.

\subsection{Ablation Study}
To investigate the impact of different training data components on DPS performance, we conduct ablation experiments by selectively removing parts of the training set. Specifically, we compare the full DPS model fine-tuned with supervised data including MS MARCO and multi-hop QA datasets, with four ablated variants: 1) \textbf{-SFT}: DPS model without supervised fine-tuning; 2) \textbf{-MS}: DPS trained without MS MARCO data; 3) \textbf{-Multihop}: DPS trained without multi-hop QA data; and 4) \textbf{-Synthetic}: excludes synthetic QA data generated for pretraining. This analysis helps quantify the impact of each training component on overall performance across multiple QA benchmarks. 

Table~\ref{tab:ablation} summarizes the results on multiple experiments. Removing supervised fine-tuning (-SFT) leads to a substantial performance drop, confirming the importance of supervised learning for effective passage selection. Excluding MS MARCO (-MS) data reduces general retrieval quality, resulting in lower scores across datasets. Similarly, removing multi-hop data (-Multihop) affects the performance on multi-hop benchmarks such as HotpotQA and MuSiQue, indicating that exposure to multi-hop reasoning examples is critical for DPS to capture inter-passage dependencies. Finally, excluding synthetic data (-Synthetic) also causes moderate degradation, indicating that large-scale weak supervision from synthetic examples enhances the model’s ability to handle long contexts and high-level queries, and provides valuable complementary signals for generalization. These results demonstrate that each training data component contributes to the final performance of DPS, and that a diverse, comprehensive training set is essential for robust and generalizable passage selection.

\subsection{Further Experiment}
Unlike many reranking methods that rely on large-scale models or intricate multi-stage pipelines—such as RankingGPT and RankZephyr, which leverages GPT-3.5 (~175B parameters) or GPT-4. DPS functions independently of the generation model and does not require fine-tuning of the answer generator. Even when paired with smaller generation models like Qwen2.5-7B, DPS delivers competitive results compared to these substantially larger systems. This highlights DPS’s advantages in efficiency, scalability, and practicality for real-world applications where model size, inference cost, and ease of integration are critical.

We further evaluate the scalability and robustness of DPS combined with various backbone LLMs of differing sizes and capabilities. As shown in Table~\ref{tab:further_exp}, DPS consistently improves performance regardless of the underlying generation model. With smaller models such as Qwen2.5-7B and Llama3.1-8B, DPS achieves substantial gains over their standalone baselines, demonstrating the effectiveness of dynamic passage selection across different model scales. Notably, DPS paired with stronger models like Qwen2.5-72B and GPT-4o attains the best overall results, achieving up to 69.02 F1 on HotpotQA and 32.40 F1 on MuSiQue. These results underscore DPS’s strong adaptability and suggest it can serve as a plug-and-play module to significantly enhance retrieval-augmented reasoning performance across a wide range of LLMs and domains.

\section{Conclusion}
We propose DPS, a novel dynamic passage selection framework for information retrieval tasks. DPS explicitly models inter-document dependencies and adaptively selects a minimal subset of passages required for answer generation. Unlike traditional reranking approaches that assign independent scores and rely on a fixed Top-\textit{K} strategy, DPS reformulates passage selection as a structured prediction task. By leveraging large language models to reason about combinatorial relevance, DPS provides a principled and adaptive alternative to rigid ranking pipelines. Extensive experiments across five diverse benchmarks—including multi-hop and domain-specific datasets—demonstrate that DPS consistently outperforms strong baselines and recent rerankers, achieving new state-of-the-art results. Notably, DPS achieves these gains without requiring generation model retraining or pipeline modifications, highlighting its practicality and scalability. 

While DPS shows strong performance across settings, several limitations remain for future research. DPS is limited by the maximum input token length of the LLM, and when the candidate passage set is too large, DPS needs to employ a sliding window strategy for inference. Moreover, he approach still depends on the initial retrieval step; missing key candidates can hinder downstream selection. Moreover, current selection does not explicitly model reasoning chains or intermediate steps, which may be necessary for more complex queries. Although DPS does not require generator fine-tuning, it still relies on supervised data to train the selector, limiting its applicability in low-resource domains. Inference cost, though moderate, is higher than pointwise rerankers, which may impact latency in time-sensitive scenarios.

Despite these limitations, we view DPS as an important step toward more adaptive, reasoning-aware retrieval in information retrieval. In future work, we plan to integrate explicit reasoning techniques to further improve evidence selection. We also aim to explore lightweight or instruction-tuned versions of DPS for zero-shot generalization, and unify retrieval and selection under a single training objective. Beyond QA, we envision DPS benefiting broader applications including citation generation, open-domain dialogue, and multimodal grounding—paving the way for more reliable and context-sensitive generation systems.

\bibliography{aaai2026}

\end{document}